\documentclass[10pt]{article} 

\usepackage[preprint]{rlj} 

\usepackage{amssymb}            
\usepackage{mathtools}          
\usepackage{mathrsfs}           
\usepackage{graphicx}           
\usepackage{subcaption}         
\usepackage[space]{grffile}     
\usepackage{url}                
\usepackage{lipsum}             
\usepackage{algorithm}
\usepackage{algpseudocode}
\newcommand{\commentalg}[1]{\hfill{\scriptsize\textcolor{gray}{#1}}}

\title{Shielded Controller Units for RL with Operational Constraints Applied to Remote Microgrids}

\setrunningtitle{Enter Your Running Title Here}

\emails{nekoeihe@mila.quebec}

\author{
    Hadi Nekoei\textsuperscript{\rm 1,2,3},
    Alexandre Blondin-Massé\textsuperscript{\rm 4},
    Rachid Hassani\textsuperscript{\rm 4},
    Sarath Chandar\textsuperscript{\rm 1,3,5,6},
    Vincent Mai\textsuperscript{\rm 4}
}
\affiliations{
    \textsuperscript{\rm 1}\,Chandar Research Lab
    \textsuperscript{\rm 2}\,Université de Montréal
    \textsuperscript{\rm 3}\,Mila – Quebec AI Institute
    \textsuperscript{\rm 4}\,IREQ – Hydro-Québec Research Institute
    \textsuperscript{\rm 5}\,Polytechnique Montréal
    \textsuperscript{\rm 6}\,CIFAR AI Chair\\
}

\contribution{
    Provide a succinct but precise list of the contribution(s) of the paper. Use contextual notes to avoid implications of contributions more significant than intended and to clarify and situate the contribution relative to prior work (see the examples below). If there is no additional context, enter ``None''. Try to keep each contribution to a single sentence, although multiple sentences are allowed when necessary. If using complete sentences, include punctuation. If using a single sentence fragment, you may omit the concluding period. A single contribution can be sufficient, and there is no limit on the number of contributions. Submissions will be judged mostly on the contributions claimed on their cover pages and the evidence provided to support them. Major contributions should not be claimed in the main text if they do not appear on the cover page. Overclaiming can lead to a submission being rejected, so it is important to have well-scoped contribution statements on the cover page.
    }
    {
    None
    }

\contribution{
    The submission template for submissions to RLJ/RLC 2025
    }
    {
    Built from previous RLC/RLJ, ICLR, and TMLR submission templates
    }

\contribution{
    \textit{{[}Example of one contribution and corresponding contextual note for the paper ``Policy gradient methods for reinforcement learning with function approximation'' \citep{Sutton2000}.]}\\ This paper presents an expression for the policy gradient when using function approximation to represent the action-value function.
    }
    {
    Prior work established expressions for the policy gradient without function approximation \citep{Williams1992}.
    }

\keywords{RLJ, RLC, formatting guide, style file, \LaTeX~template.} 

\summary{The summary appears on the cover page. Although it can be identical to the abstract, it does not have to be. One might choose to omit the stated contributions in the Summary, given that they will be stated in the box below. The original abstract may also be extended to two paragraphs. The authors should ensure that the contents of the cover page fit entirely on a single page. The cover page does \textbf{not} count towards the 8--12 page limit.

\lipsum[1]
}

\begin{document}

\maketitle  


\begin{abstract}
Reinforcement learning (RL) is a powerful framework for optimizing decision-making in complex systems under uncertainty—an essential challenge in real-world settings, particularly in the context of the energy transition. A representative example is remote microgrids that supply power to communities disconnected from the main grid. Enabling the energy transition in such systems requires coordinated control of renewable sources like wind turbines, alongside fuel generators and batteries, to meet demand while minimizing fuel consumption and battery degradation under exogenous and intermittent load and wind conditions.
These systems must often conform to extensive regulations and complex operational constraints. To ensure that RL agents respect these constraints, it is crucial to provide interpretable guarantees. In this paper, we introduce Shielded Controller Units (SCUs), a systematic and interpretable approach that leverages prior knowledge of system dynamics to ensure constraint satisfaction. Our shield synthesis methodology, designed for real-world deployment, decomposes the environment into a hierarchical structure where each SCU explicitly manages a subset of constraints.
We demonstrate the effectiveness of SCUs on a remote microgrid optimization task with strict operational requirements. The RL agent, equipped with SCUs, achieves a 24\% reduction in fuel consumption without increasing battery degradation, outperforming other baselines while satisfying all constraints. We hope SCUs contribute to the safe application of RL to the many decision-making challenges linked to the energy transition. All code and supporting data are available at \url{https://github.com/chandar-lab/SCU_RL_MicroGrid}.
\end{abstract}


\section{Introduction}
\label{sec:introduction}

Ensuring safe, sustainable, and reliable operation is critical for the control of real-world systems, especially in the context of the energy transition. These systems, such as remote microgrids serving communities without access to the main power grid, must adhere to numerous norms and regulations. Any algorithm deployed on such a system must have its compliance to these constraints explicitly and interpretably proven. The challenge to implement robust and provable compliance guarantees for reinforcement learning (RL) agents hinders their widespread use in critical industrial settings , despite RL's potential for solving complex optimization problems in real-world  applications \citep{Dulac-Arnold2019, Kober2013}. Addressing this issue is essential for unlocking RL's full potential in industrial settings, where a wrong decision could lead to equipment damage, system failures, safety incidents, or significant economic losses \citep{Garcia2015, Amodei2016}.

White-box shielding is a promising approach to meet these requirements \citep{Alshiekh2018, Jansen2018}. Synthesized from prior knowledge of environment dynamics, a white-box shield verifies that the agent's action do not violate constraints, and imposes an alternative complying action when necessary. This mechanism, produced in human-understandable logic, can be audited by system experts, and adjusted to the desired level of conservativeness to handle the system's uncertainties. Significant progress has been made in shield synthesis \citep{Könighofer2020, Bloem_2020} and white box shielding \citep{Hsu_Hu_Fisac_2023, Krasowski_Thumm_Muller_Schafer_Wang_Althoff_2023}. However, in complex systems with many constraints across multiple devices and varying time horizons, developing a provable shield in a systematic way remains challenging. Indeed, shielding approaches usually require expressing constraints logically, simulating environment rollouts, and designing reasonably performing fallback policies. Doing so for complex environments can be tedious and error prone. A methodology decomposing the shielding problem into simpler sub-problems and allowing compliance to be proved independently for each constraint would considerably facilitate this process, enhancing the applicability of RL in critical settings.

In this paper, we propose Shielded Controllers Units (SCUs) for RL, a systematic methodology aimed at this objective. The environment is decomposed into SCUs, which link a controller to a real system to ensure the compliance of a corresponding set of constraints. Systems can be devices or a group of lower-level SCUs, forming a hierarchical structure. A controllers consists of a shielded action dispatcher and a digital twin of its corresponding system. The digital twin combines a state estimator, updated from sensor measurements, and a simulator to predict future system behavior. The shielded action dispatcher uses prior knowledge and the digital twin to test the action's validity before relaying it to the real system. Each SCU possesses a shield to guarantee compliance with its associated set of constraints. The SCU structure ensures full compliance over the whole system, while considerably simplifying the shield design process for complex real-world environment.

We demonstrate SCUs on the problem of remote microgrid optimization. Remote microgrids are small power systems disconnected from the main grid, typically powered by fossil fuel generators (\emph{gensets}). To reduce CO2 emissions and operational costs, electricity utilities are installing wind turbines for fossil-free energy. Due to the intermittent nature of wind power, large capacity batteries are needed for flexibility, while gensets remain essential to reliably meet power demand. The power dispatch problem aims to optimize the use of batteries and gensets to minimize fuel consumption and battery degradation. This complex sequential decision-making problem involves long-term planning, uncertain exogenous variables such as power demand and wind, and many constraints to ensure reliable and sustainable operation. When modeled in a realistic industrial scenario, it is a difficult problem for non-learning based optimization methods, making it an excellent use case to demonstrate the power of the shielded controllers: our RL agent outperforms baselines while its compliance to every constraint is guaranteed.
This paper makes three contributions:
\begin{itemize}
    \item It introduces Shielded Controller Units (SCUs), a systematic approach to white-box shield design, improving interpretability while ensuring constraint compliance for RL agents in industrial contexts.
    \item It presents a realistic remote microgrid optimization problem under uncertainty, aimed at training and testing an RL agent with real-world operational constraints.
    \item The SCU approach is applied to microgrid optimization, ensuring constraint compliance while the resulting RL agent outperforms current baselines for this problem. 
\end{itemize}

\section{Related work}
\label{sec:related_work}


\textbf{Constrained RL} seeks to optimize policies under safety constraints and is an active area of research~\citep{wachi2024surveyconstraintformulationssafe, ijcai2021p614}. Black-box methods like CPO~\citep{achiam2017constrainedpolicyoptimization} and CVPO~\citep{liu2022constrainedvariationalpolicyoptimization} rely on cost signals to discourage unsafe behavior but often lack interpretability and struggle with learning~\citep{mani2025risk}. Shielding approaches that train classifiers to block unsafe actions~\citep{waga2022dynamicshieldingreinforcementlearning} face similar issues and depend heavily on environmental knowledge.
White-box shielding instead leverages known dynamics. Early work~\citep{bloem2012synthesis, Alshiekh2018} used formal methods to synthesize shields, while later efforts incorporated runtime monitoring~\citep{Fulton_Platzer_2018}, probabilistic safety constraints~\citep{Bouton_Karlsson_Nakhaei_Fujimura_Kochenderfer_Tumova_2019}, and adaptive interventions via model predictive control~\citep{Banerjee_Rahmani_Biswas_Dillig_2024}. Reviews~\citep{Krasowski_Thumm_Muller_Schafer_Wang_Althoff_2023, Hsu_Hu_Fisac_2023} summarize the tradeoffs between black-box and model-based approaches.
However, these approaches can be difficult to deploy in practice. They often require formal logic, simulated rollouts, and/or fallback policies, which are tedious and error-prone to design in complex systems. We address this by introducing a practical, modular methodology that decomposes safety enforcement into simpler shielding sub-problems, making shield design easier.
The closest to our work is factored shielding (ElSayed-Aly et al. 2021), which improves scalability in multi-agent RL by assigning localized shields to subsets of agents equivalent to horizontal decomposition across agents. Our SCU-based approach instead leverages hierarchical decomposition of the environment and digital twins for modular, interpretable shielding for complex real-world systems.

\textbf{Microgrid optimization} aims to manage energy sources for reliable, low-emission, and sustainable operation, accounting for battery degradation. This requires long-term planning under uncertainty (e.g., power demand, wind), which challenges traditional methods due to mixed variables and non-linear dynamics. Mixed-integer programming has had some success~\citep{Hajimiragha2013, Silvente2015, Hirwa2022}, but often oversimplifies constraints and ignores battery depletion. A more realistic model~\citep{Lambert2023} supports real-time control of gensets but excludes batteries and wind.

RL-based approaches show promise~\citep{Dimeas2010, Foruzan2018, Arwa2020, Yang2021, Wang2024}, though often lack industrial realism or safety guarantees. \citet{Eichelbeck_Markgraf_Althoff_2022} apply RL for energy dispatch under similar constraints, using an action projection shield, though replacing fuel genset dynamics with islanding robustness. \citet{Ceusters_Camargo_Franke_Nowé_Messagie_2023} train an RL agent with a physics-based safety layer and fallback policy for a multi-energy system. Other constrained RL work spans related domains: cooling~\citep{Yu_Zhang_Song_Hui_Chen_2024}, EV charging~\citep{Zhang_Guan_Che_Shahidehpour_2024}, and voltage regulation~\citep{Chen_Shi_Arnold_Peisert_2021}.

\section{Shielded controller units}
\label{sec:shielded_controller_units}

In complex, critical environments like real-world industrial problems, numerous constraints must be respected across systems with multiple devices. Some constraints pertain to individual devices, while other concern groups of devices, binding their control together. Ensuring compliance with a constraint sometimes requires considering future environment states to guarantee that an action is recoverable. To satisfy operational norms, any agent deployed on such complex system must interpretably ensure compliance with these constraints, despite uncertainties from exogenous variables.

\begin{figure*}[!h]
    \centering
    \begin{subfigure}{0.27\textwidth}
        \centering
        \includegraphics[width=1\linewidth]{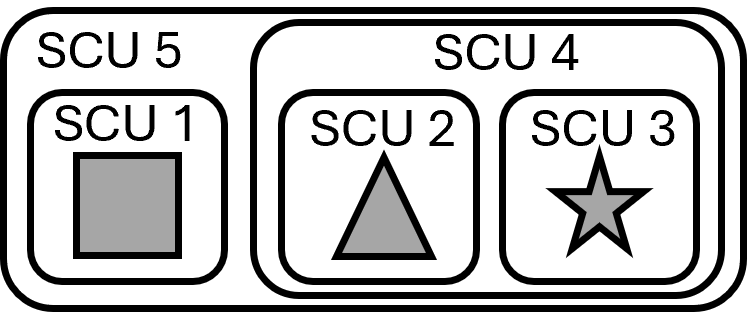}
        \caption{Example of SCU hierarchy. Shapes represent devices in a complex environment.}
        \label{fig:SCU_Example}
    \end{subfigure}
    \hfill
    \begin{subfigure}{0.7\textwidth}
        \centering
        \includegraphics[width=0.48\linewidth]{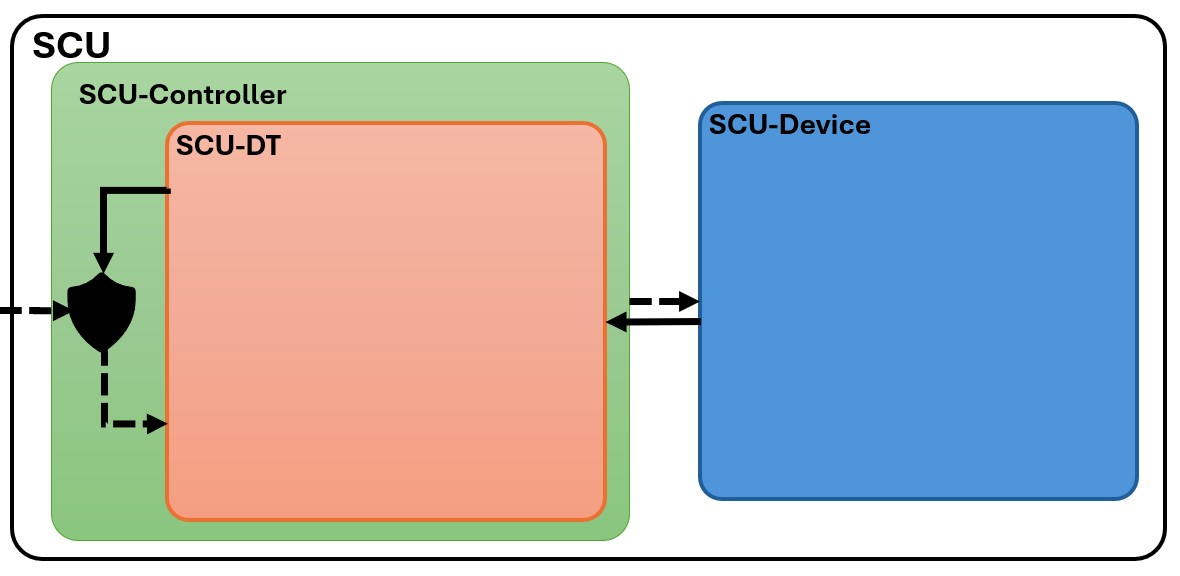}
        \includegraphics[width=0.48\linewidth]{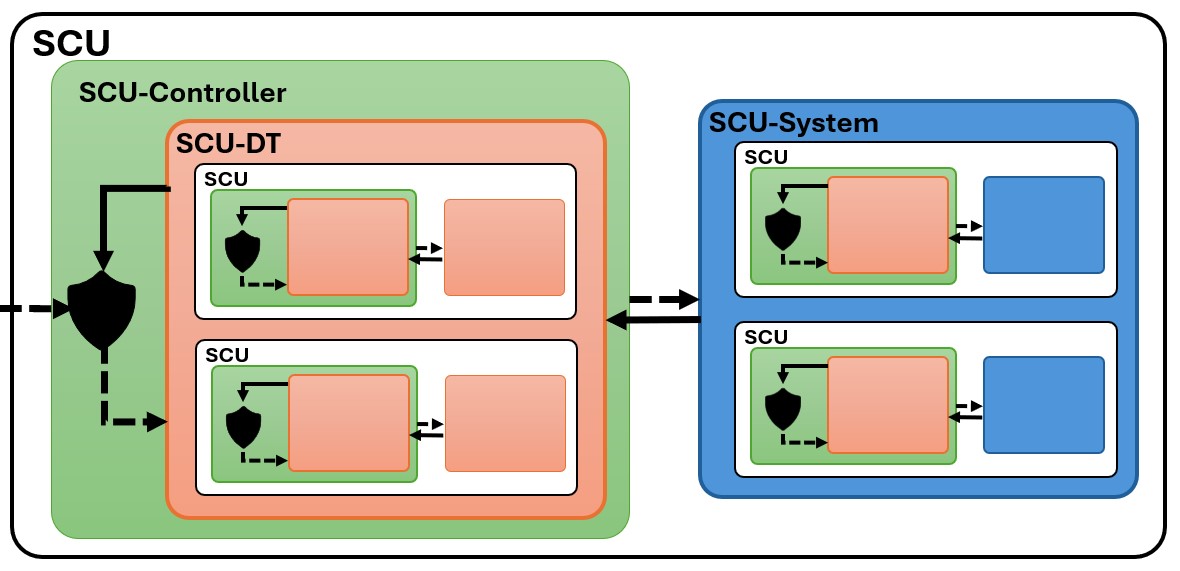}
        \caption{General SCU structure. Left: SCU for a single device. Right: higher-level SCU for a composite system.}
        \label{fig:shieldedcontrollerunit}
    \end{subfigure}
    \caption{Illustrations of Shielded Controller Units (SCUs) at different levels of abstraction.}
    \label{fig:combined_scu}
    \vspace{-5mm}
\end{figure*}

Our proposed methodology addresses this problem systematically by decomposing the environment into Shielded Controller Units (SCUs). This decomposition is based on analyzing the set of constraints. Every constraint pertains to a subset of the environment's devices, and constraints corresponding to the same subset are implemented in the same SCU. SCUs are implemented hierarchically, meaning they cannot overlap without one containing the other. For example, consider a system composed of square, triangle and star devices as shown in Figure \ref{fig:SCU_Example}. Each devices has its own constraints, and thus its own SCU: 1, 2 and 3. Two additional constraints might concern the triangle and the star: they would be managed by SCU 4. If another constraint links the triangle and square devices, whether or not the star is involved, it cannot be treated outside of SCU 4. Therefore, SCU 5 will contain both SCU 1 and SCU 4. 

\subsection{Elements of shielded controller units}

An SCU links a shielded controller to a real system. The controller receives an action from a higher-level SCU, ensures it respects constraints at its level, and relays it to the real system. It contains two elements: a digital twin and a shielded dispatcher. The shielded dispatcher enforces constraints by checking each action and sending a corrected one if needed for every system subcomponent. Many white-box shielding methods exist; choices depend on constraint type and prior knowledge. Some rely on state estimation~\citep{Carr_Jansen_Junges_Topcu_2022} or model predictive shielding~\citep{bastani2020safereinforcementlearningnonlinear, Banerjee_Rahmani_Biswas_Dillig_2024}. 
This need is met by the digital twin, combining a state estimator updating internal state from sensors and a simulator predicting system behavior from the current state and action sequence. Importantly, the digital twin must simulate all SCUs of components to ensure compliance with constraints. This includes the digital twin for the real system \textit{and} the digital twin inside the SCU controller, as shown in Figure~\ref{fig:shieldedcontrollerunit}.

\subsection{Step function and interactions}

The SCU structure operates hierarchically, processing the action sent by the agent from the top level SCU down to the lower levels, and propagating the information from the devices sensors back up to the agent while updating the various levels of digital twins. This process can be realized with a single step function called from the top SCU, as described in algorithm \ref{alg:scu_step}. It is divided in three substeps.
\begin{enumerate}[leftmargin=0pt, labelwidth=0pt, align=left, nosep]
    \item First, at line 3, the shield dispatcher receives action $a_t$ from the higher level and dispatches it into complying actions $\{a_t^{\mathrm{comp}, i}\}_{i = 1\dots k}$ 
    using function $\textsc{Shield}$. This function is designed ad-hoc via an interpretable shielding method, potentially with digital twin $U^{\mathrm{DT}}$.
    \item Next, lines 4 to 11, the safe actions are propagated to the real system's \textsc{Step} function. If it is a device, it returns observation $o_{t+1}$. If it is a system of SCUs, it will return set $\{\hat{s}^i_{t+1}\}_{i=1\dots k}$ of state estimations of the subcomponents' SCUs. These are aggregated in observation $o_{t+1}$.
    \item Finally, at line 12, the controller internal state $s^\mathrm{SC}_{t+1}$ and its numerical twin's  $s^\mathrm{SC\cdot DT}_{t+1}$ are updated using function \textsc{UpdateController} in algorithm \ref{alg:controller_update}. $s^\mathrm{SC}_{t+1}$ is updated with ad-hoc function \textsc{ControllerState} and $s^\mathrm{SC\cdot DT}_{t+1}$ with function \textsc{UpdateDT}. If the SCU is of a device, \textsc{UpdateDT} estimates $s^\mathrm{DT}_{t+1}$ using ad-hoc estimation function \textsc{StateEstim}. If however the SCU has $k$ SCUs as sub-components $U^j$, there are 2 elements to update per $U^j$: (1) digital twin $U^{j, \mathrm{DT}}$ of the real device and (2) controller $U^{j, \mathrm{SC}}$ and its internal digital twin $U^{j, \mathrm{SC\cdot DT}}$. This implies a new recursive function propagating $o_{t+1}$ to lower levels $j$. It is disaggregated into $o^j_{t+1}$ which are used for \textsc{UpdateDT} on $U^{j, \mathrm{DT}}$ and \textsc{UpdateController} on $U^{j, \mathrm{SC}}$. Finally, the digital twin new state $s^\mathrm{DT}_{t+1}$ is the aggrefation of lower level SCU states $s^j_{t+1}$.

\end{enumerate}

\vspace{-2mm}
\begin{algorithm}
\small
\caption{SCU step function}\label{alg:scu_step}
\begin{algorithmic}[1]
\Function{Step}{$U$, $a_t$ } \commentalg{$U$: SCU, $a_t$: action}
\State $k \gets U.\Call{NbComponents}{}$ 
\State $\{a^{\mathrm{comp}, i}_t\}_{i=1\dots k} \gets \Call{Shield}{U^{\mathrm{DT}}, a_t}$
\If{$U.\Call{IsDevice}{}$} \commentalg{$k = 1$}
\State $o_{t+1} \gets$ \Call{Step}{$U^{\mathrm{dev}}, a^{\mathrm{comp}, 1}_t$}
\Else \;\commentalg{$k > 1$}
    \For{$i \in \{1 \dots k\}$}
        \State $s^{i}_{t+1} \gets \Call{Step}{U^i, a^{\mathrm{comp}, i}_t}$
    \EndFor
    \State $o_{t+1} \gets \Call{Obs}{\textsc{Agg}(\{s^i_{t+1}\}_{i=1\dots k})}$
\EndIf
\State $s_{t+1} \gets \Call{UpdateController}{U^{\mathrm{SC}}, o_{t+1}}$
\State \Return $s_{t+1}^\mathrm{DT}$
\EndFunction
\end{algorithmic}
\end{algorithm}
\vspace{-9mm}
\begin{algorithm}[H]
\small
\caption{Controller update function}\label{alg:controller_update}
\begin{algorithmic}[1]
\Function{UpdateController}{$U^\mathrm{SC}, o_{t+1}$}
\State $s^\mathrm{SC}_{t+1} \gets \text{ControllerState}(U^\mathrm{SC}, o_{t+1})$
\State $s^\mathrm{SC\cdot DT}_{t+1} \gets \text{UpdateDT}(U^\mathrm{SC\cdot DT}, o_{t+1})$
\State $s_{t+1} = [s^\mathrm{SC}_{t+1}, s^\mathrm{SC\cdot DT}_{t+1}]$
\State \Return $s_{t+1}$
\EndFunction
\end{algorithmic}
\end{algorithm}

\vspace{-9mm}

\begin{algorithm}[H]
\small
\caption{Digital twin update function}\label{alg:dt_update}
\begin{algorithmic}[1]
\Function{UpdateDT}{$U^\mathrm{DT}, o_{t+1}$}
\If{$U.\Call{IsDevice}{}$}
    \State $s^\mathrm{DT}_{t+1} \gets \text{StateEstim}(s^\mathrm{DT}_{t}, o_{t+1})$
\Else
    \State $k \gets U.\Call{NbComponents}{}$ 
    \State $\{o^j_{t+1}\}_{j=1\dots k}\gets \text{Disagg}(o_{t+1})$
    \For{$j \in 1 \dots k$} 
    \State $s^{j,\mathrm{SC\cdot DT}}_{t+1} \gets \text{UpdateDT}(U^{j, \mathrm{DT}}, o^j_{t+1})$ 
    \State $s^j_{t+1} \gets \text{UpdateController}(U^{j,\mathrm{SC}}, o^j_{t+1})$
    \EndFor
    \State $s^\mathrm{DT}_{t+1} \gets \text{Agg}(s^j_{t+1})$
    \EndIf
    \State \Return $s^\mathrm{DT}_{t+1}$
\EndFunction
\end{algorithmic}
\end{algorithm}

\subsection{Integration within the RL framework}
The RL agent interacts with the environment using the usual Markov Decision Process (MDP) framework, but only through the highest level SCUs' step functions. These receive their respective parts of the action space's inputs and propagate them to their subcomponents down to the devices. The agent's observation is the aggregation of high-level SCUs' controllers' state estimation.

\section{Shielded remote microgrid optimization}
\label{sec:remote_microgrid_optimization}

\begin{figure}[!t]
    \centering
    \includegraphics[width=1\linewidth]{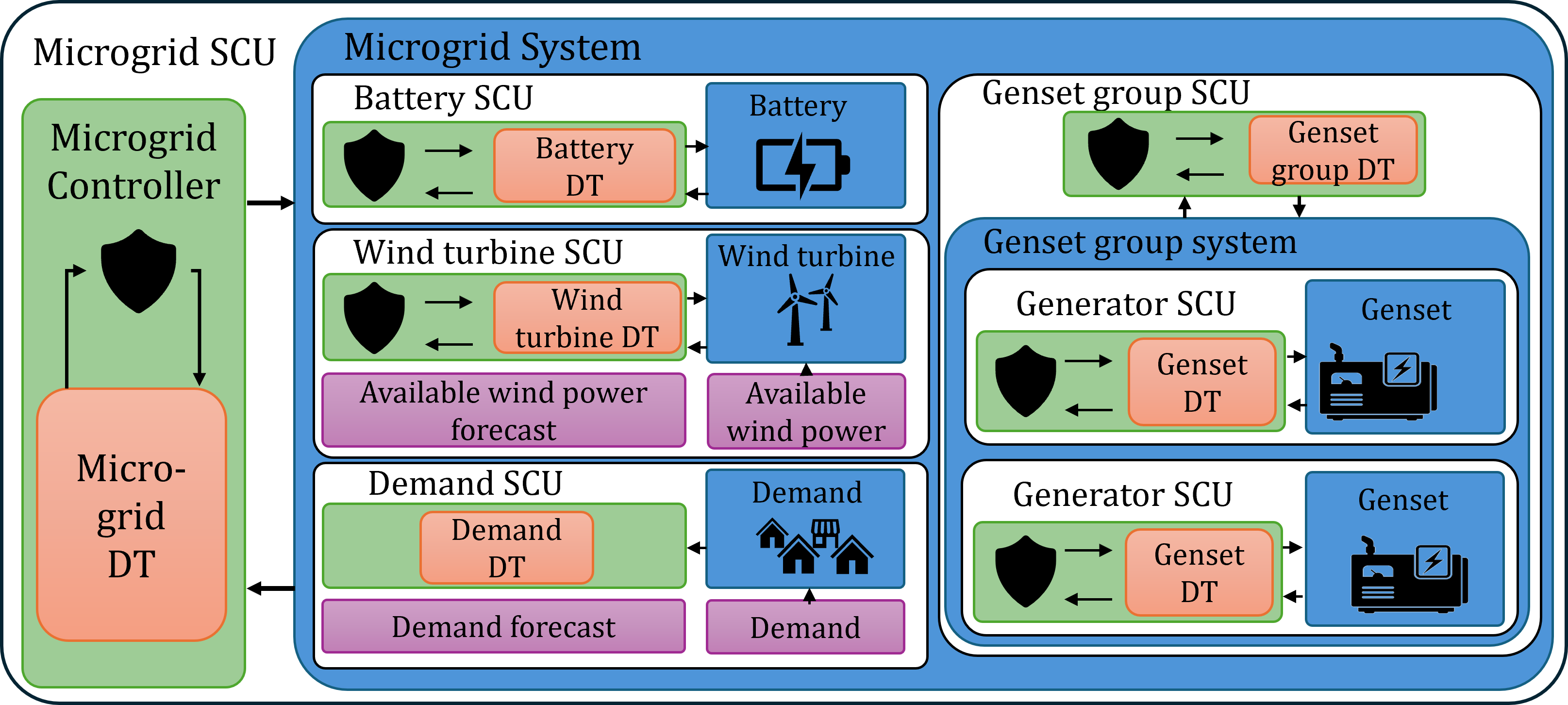}    
    \caption{\em SCU decomposition of the microgrid.}
\label{fig:microgrid_scu}
\vspace{-5mm}
\end{figure}

We consider a remote microgrid composed of a wind turbine, a battery and two fuel gensets, generating electric power to meet the demand of a village. The objective of the power dispatch problem is to control these four devices to minimize genset fuel consumption and battery depreciation, while ensuring that the generated power satisfies the demand at every time step. In this problem, two variables are considered exogenous and beyond the agent's control: demand and available wind power. Specialized industry models can provide predictions to the agent, but they are uncertain.
This problem, when realistically modeled to target industrial deployment, is an excellent test case for the SCU approach. It is of reasonable scale for in-depth analysis but complex enough that non-learning based methods struggle with real time control. RL is a powerful alternative, but it must provably respect the industrial constraints. 
In this section, we describe how we modeled the environment's physical dynamics, rewards and constraints. We also discuss how the environment is decomposed in SCUs (See Figure~\ref{fig:microgrid_scu}), the shielding approach for each controller, and define the MDP the RL agent actually interacts with.

\subsection{Exogenous variables}
There are two exogenous variables in the environment, over which the agent has no control: the power demand $\delta_t$ and the available wind power $p^\mathrm{wind, avail}_t$. We use real data gathered from remote microgrids over one year. 
Demand ranges from 180 to 540 kW, averaging 320 kW, while available wind power ranges from 0 to 400 kW, averaging 272 kW. 
The agent observes both variables at every time step and receives realistic predictions over a 30-point forecast at 15 minutes intervals, covering 7.5 hours. These predictions were generated using industry-standard approaches. Demand forecasts were produced using a general additive model (GAM) that combines transformations of the consumption observed the previous day and the previous week, along with observed and forecast temperatures \citep{demand-forecast}. 
Wind power forecasts were generated by downscaling meteorological variables and converting them using an 8-direction power curve model. Forecasts were then fine-tuned using turbine-specific loss models, availability, and a statistical auto-adaptive model with real-time data.

\subsection{Devices}
The modeled microgrid includes three types of device: one wind turbine, two gensets, and one battery. Each device is managed by an SCU. The industrial norms they must comply with are described in \textbf{bold}. In this experiment, the digital twin sensor update function for devices and systems has been implemented as exact state update, which is realistic given the availability of precise sensors to measure relevant state variables for batteries, gensets and wind turbines.

\paragraph{Wind turbine}
The wind turbine generates renewable electric power for the microgrid by harnessing wind energy at no cost. At step $t$, it receives action $a_t = \Bar{p}^{\mathrm{wind}}_t$ containing the wind power curtailment setpoint. The resulting power $p^{\mathrm{wind}}_t$ is $\Bar{p}^{\mathrm{wind}}_t$ clipped between 0 and current available wind power $p^\mathrm{wind, avail}_t$. In this power dispatch problem, \textbf{no operational constraints applies} on the wind turbine, so the shield directly relays $a^{\mathrm{comp}}_t = a_t$ to the device.

\paragraph{Battery}
The battery stores chemical energy and can absorb or release electrical power $p^{\mathrm{batt}}_t$. The power $p^{\mathrm{batt}}_t$ responds to the $\Bar{p}^{\mathrm{batt}}_t$ setpoint from action $a_t$, can be positive (discharge) or negative (charge), and is limited by the battery nominal power of 600 kW. The battery state of charge (SoC) indicates the accumulated energy as a percentage of the battery 672 kWh capacity. Charging and discharging are inefficient processes, leading to a 5\% energy loss in each direction. We model the battery's physical behavior precisely according to \cite{G-2024-49}. The battery's degradation over charging/discharging cycles $d_{b,t}$ is modeled in arbitrary units using an approximated online rainflow algorithm adapted for MDPs, inspired from \cite{Obermayr_Riess_Wilde_2021} and \cite{Kwon_Zhu_2022} and detailed in the appendix.  To prevent damage to the battery, the \textbf{SoC is constrained between 90 and 10\% of capacity}, or 5\% as a reserve in case of emergency. The shield uses prior knowledge of the battery inner dynamics to determine the corresponding limits and accordingly clips setpoint $\Bar{p}^{\mathrm{batt}}_t$ received from $a_t$ before relaying it as $a^{\mathrm{comp}}_t$ to the real battery. 

\paragraph{Gensets}
Gensets can produce electric power from fuel energy. We model their behavior and constraints based on \cite{Lambert2023}. Gensets receive action $a_t = (\Delta^{\mathrm{gen}}_t, \Bar{p}^{\mathrm{gen}}_t)$. Gensets's running status can be \textit{on} or \textit{off}, changed by \textit{start} or \textit{stop} command $\Delta^{\mathrm{gen}}_t$. Once \textit{on}, their power production $p^{\mathrm{gen}}_t$ is controlled by setpoint $\Bar{p}^{\mathrm{gen}}_t$ ranging between 0 and maximum power of 440 kW. Fuel consumption is proportional to $p^{\mathrm{gen}}_t$ at 0.25 l/kWh, plus a constant 10 l/h when the genset is \textit{on}.

To prevent long term damage, several operational constraints must be respected. When changing status, \textbf{gensets must complete routines}, like a 3-minute warm-up producing 100 kW or a 5-minute cool-down producing 0 kW. Once on, they have \textbf{minimum runtime} of 30 minutes during which they should not be turned \textit{off}. Except during these routines, \textit{on} gensets have a \textbf{minimal power production} of 120 kW. Producing the maximum 440 kW continuously can damage the device: \textbf{it should instead be capped at its nominal capacity} of 400 kW, except in emergencies where the 40 kW overload reserve can be used. Finally, to meet ISO 8528-1 standards \citep{iso8528-1-2018}, the \textbf{average power over the last 48 hours of operation is capped} at 70\% of nominal capacity.  
The genset SCU controller ensures these constraints are respected by modifying violating $\Delta^{\mathrm{gen}}_t$ and clipping $\Bar{p}^{\mathrm{gen}}_t$ from $a_t$ before relaying them as $a^{\mathrm{comp}}_t$ to the physical genset.

\subsection{Systems}

Two sets of operational constraints for the microgrid concern specific groups of devices, leading to two additional SCUs: the genset orchestrator and the microgrid.

\paragraph{Genset orchestrator} 
The genset orchestrator manages the two gensets, its subcomponents, by dispatching action $a_t = (\Delta^{\mathrm{orch}}_t, \Bar{p}^{\mathrm{orch}}_t)$ into actions $a^{\mathrm{comp}, i}_t = (\Delta^{\mathrm{gen}, i}_t, \Bar{p}^{\mathrm{gen}, i}_t)$ complying with two operational constraints. First, genset $i$ cannot be turned \textit{on} if genset $i - 1$ is turned \textit{off}, and vice versa. This \textbf{priority order constraint} ensures compatibility with existing heuristic-based systems so they can take over if necessary, as described in \cite{G-2024-49}. The shielded dispatcher follows a state-machine logic allowing it to turn \textit{on} or \textit{off} only one specific generator at the time. This allows to dispatch $a_t$'s single status change command $\Delta^{\mathrm{orch}}_t$ to \textit{start} or \textit{stop} a generator, or keep the current configuration, which the shield defaults to if $\Delta^{\mathrm{orch}}_t$ is inapplicable. 

The other constraint is the \textbf{equal power fraction} presented in \cite{Lambert2023}: all running genset not in warmup or cooldown must run at the same percentage of their nominal powers. The orchestrator uses its digital twin to account for single genset constraints, such as the 70\% maximum average power over 48h, as it may limit both gensets together. This constraint binds all $\Bar{p}^{\mathrm{gen}, i}_t$ together: the orchestrator dispatches them in $a^{\mathrm{comp}, i}_t$ so that the total power produced by the gensets, $p^{\mathrm{orch}}_t$, is as close as possible to the orchestrator setpoint $\Bar{p}^{\mathrm{orch}}_t$ received in $a_t$.

\paragraph{Microgrid}
The microgrid is the highest level SCU. Its objective is to ensure that \textbf{the balance, i.e. the difference between power generation and demand, is kept at 0 for all time steps}. A negative balance, or shortage,  is a critical failure of the system, while a positive balance should also be avoided although it is easier to recover from. To enforce this constraint, the microgrid subcomponents are the battery SCU, the wind turbine SCU, and the genset orchestrator SCU. Each time step, the microgrid shield dispatches complying actions $a^{\mathrm{comp}, i}_t$ to its three subcomponents, as power setpoints $\Bar{p}^{\mathrm{batt}}_t$,  $\Bar{p}^{\mathrm{wind}}_t$ and $\Bar{p}^{\mathrm{orch}}_t$, and orchestrator status change $\Delta^{\mathrm{orch}}_t$. The balance constraint, and a hard-coded wind over fuel priority, reduce the degrees of liberty of the received action $a_t$ to $\Delta^{\mathrm{orch}}_t$ and  $\Bar{p}^{\mathrm{batt}}_t$.

To ensure 0 balance, shielding operates at two levels. At any given time step $t$, the subcomponents setpoints must ensure the sum of generated powers $p^{\mathrm{batt}}_t$,  $p^{\mathrm{wind}}_t$ and $p^{\mathrm{orch}}_t$ meets demand $\delta_t$. It priorizes respecting the agent's battery setpoint, then using wind before fuel. If the demand is not met due to the limits of wind and gensets, $\Bar{p}^{\mathrm{batt}}_t$ is adjusted. If all the wind power cannot be absorbed by the demand and the battery, it is curtailed. To ensure that every subcomponent constraint is accounted for, the microgrid digital twin simulates each SCU's response to given commands. 

The shield must also avoid actions leading to irrecoverable future states, characterized by the genset orchestrator's inability to provide the power to balance the demand due to a genset being in cool-down/warm-up routine. Such blockages can take a maximum of 8 minutes.
To prevent this, a \textit{recovery shield} using a predictive shielding approach is deployed \citep{bastani2020safereinforcementlearningnonlinear, Hsu_Hu_Fisac_2023}. Monitoring is done by simulating the environment with the digital twin over a 9-minute period, given a \textit{fallback} genset orchestrator policy $\{\Delta^{\mathrm{orch}}_t, \Delta^{\mathrm{orch}}_{t+1}, \dots , \Delta^{\mathrm{orch}}_{t+8}\}$ with future $\Delta^{\mathrm{orch}}_{t+\tau} =  \mathrm{Start}$ for $\tau \in 1, \dots, 8$.  There are three unknown variables when simulating the next steps: exogenous demand, available wind power, and agent battery commands. The recovery shield guarantee should hold irrespective of the precision of exogenous variable predictions, so they are not used here. Instead, two scenarios are considered: (1) The worst case scenario with reserve, where battery command is set to discharge to the maximum, and demand and wind are simulated as rising and decreasing, respectively, towards their historical high and low at their fastest historical rate of change. The genset and battery devices are simulated as able to access their emergency reserve power. (2) A constant scenario for wind and demand, but where simulated gensets and batteries are not allowed to access their reserves.
Scenario (1) ensures a strong guarantee on the 0-balance constraint, while the (2) reduces reliance on reserve. If a negative balance is detected during one of these simulations, $\Delta^{\mathrm{orch}}_t$ is considered as possibly non-compliant and is replaced by the least different recoverable option.

\subsection{Markov Decision Process for the agent}
As the microgrid SCU is at the highest level, it interfaces with the RL agent. The agent's action space is two-dimensional: a discrete value for $\Delta^{\mathrm{orch}}_t$ and a continuous value for $\Bar{p}^{\mathrm{batt}}_t$. Its observation space includes the state estimation of the microgrid subcomponents and two time series of exogenous variable predictions -- wind and demand. 
The reward is a negative cost combining the gensets fuel consumption $\mathbf{F}_{o,t}$ (in liters) and the battery degradation $d_{b,t}$ in arbitrary units: $r_t = - (\mathbf{F}_{o,t} + \alpha d_{b,t})$, where weight $\alpha$ can be adjusted based on the utility's objective. Between equally performing solutions in terms of cost, a controller minimizing reserve usage of batteries and gensets is preferred.

\section{Experimental results}
\label{sec:experimental_results}

\subsection{Experimental setup}

We used the Soft-Actor Critic (SAC) RL algorithm \citep{haarnoja18}, which is known for its robust performance in continuous domains, to train our agents. Although we also trained the RL agents using PPO \citep{schulman2017proximalpolicyoptimizationalgorithms}, we found that SAC consistently outperformed PPO in this setting. Therefore, we report the results for the SAC agents. To process the available wind power and demand prediction time series, the agent's architecture integrates LSTMs. Details about the agent's architecture can be found in the appendix.
The agents were trained on $10^5$ one-minute environment steps divided in one-day episodes. The episodes initialization involved sampling a random time during the year and the gensets and battery initial states. We explored several hyper-parameter configurations (detailed in the appendix) as well as 10 training seeds and three values of $\alpha$ for the SAC agent. Both the baselines and the trained agents were then tested in a fixed test environment consisting of the first 10-day periods of each month from February to November to represent varying conditions. To report the results, we averaged the performance across these 100 days. For the RL agent's performance, we also averaged the performance and calculated the standard errors across the 10 training seeds. 

Note that all agents use shields during evaluation unless stated otherwise. We verified that none of the constraints are violated to confirm the validity of the shield.

\begin{figure}[!h]
\vspace{-5mm}
    \centering
    \includegraphics[width=0.9\linewidth]{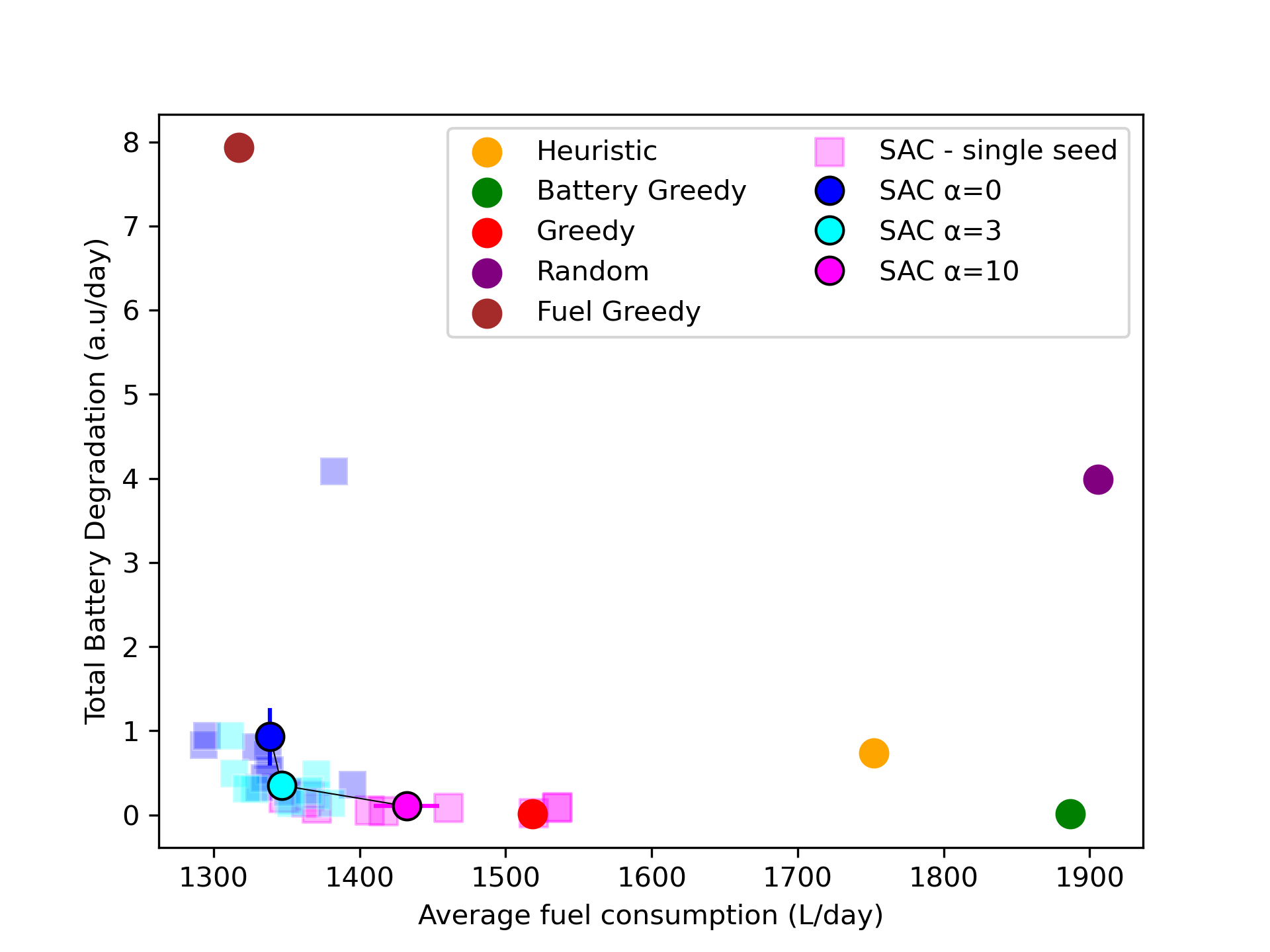} 
    \caption{\em The RL agent seeks to strike a balance between fuel consumption and battery degradation costs. Changing their respective importance in the reward function with $\alpha$ seems to exhibit the Pareto frontier. In contrast, all other baselines exhibit suboptimal performance in one or both metrics.}
    \label{fig:pareto_frontier_without_shield}

\end{figure}


\subsection{Results}

We designed our experimental section to address three research questions: (1) How do shielded RL agents perform compared to baseline algorithms in terms of minimizing fuel consumption and battery degradation? (2) How does the recovery shield impact the performance and compliance to constraints? (3) Given a shielded deployment environment, how to constrain an RL agent during training? 


\subsubsection{Performance analysis} Figure~\ref{fig:pareto_frontier_without_shield} compares the performance of our RL agents with four control baselines providing action $a_t = (\Delta^{\mathrm{orch}}_t, \Bar{p}^{\mathrm{batt}}_t)$ based on predetermined policies. 
The \textit{random} agent uniformly samples discrete $\Delta^{\mathrm{orch}}_t$ from (\textit{start, stop, do nothing}), and continuous $\Bar{p}^{\mathrm{batt}}_t$ from maximum discharging to maximum charging power. It performs poorly, showing that relying on the SCU shield is not enough to guarantee good performance.
The \textit{battery greedy}  agent lets the shield change the genset status and keeps the battery idle, with $\Delta^{\mathrm{orch}}_t$ to \textit{do nothing}, and $\Bar{p}^{\mathrm{batt}}_t$ to 0. The battery does not degrade but fuel consumption is not optimal. 
The \textit{greedy} agent tries to turn the genset off with $\Delta^{\mathrm{orch}}_t$ at \textit{stop}, and discharges the battery with maximal $\Bar{p}^{\mathrm{batt}}_t$. It consumes little fuel but highly degrades the battery. The \textit{fuel-greedy} agent also tries to turn the genset off but charges the battery when available wind power is higher than the demand, and discharges otherwise.

The industry \textit{heuristics} reproduces a policy currently deployed in industrial settings. It always keeps one genset \textit{on}, and decides to turn the other genset \textit{on} when current genset is at more than 90\% of its available power (nominal or capped by the moving average constraint) for 5 minutes in a row, or directly if it reaches more than 100\%.
It turns the second genset \textit{off} if the current total genset orchestrator power could be managed by 70\% of the other genset's available power for 5 minutes in a row.
On the battery side, it switches between two modes: charging, or discharging, to avoid fast cycling. In charging mode, it only charges  $\Bar{p}^{\mathrm{batt}}_t$ to capture an eventual excess power from the wind. It does not use the genset power to charge. Discharge only happens in case of emergency, as enforced by the shield. The battery charges until achieving 90\% SoC, and then switches to discharging mode. In discharging mode,  $\Bar{p}^{\mathrm{batt}}_t$ is set to maximum discharging power until the battery reaches 10\% SoC. It then switches back to charging mode. Performance-wise, it provides a relevant comparison point in terms of trade-off between both metrics.

The RL agents trained with SAC perform significantly better, improving fuel efficiency by 24\% for similar battery degradation.  
Modifying parameter $\alpha$ to balance battery degradation cost and fuel consumption in the reward function seems to exhibit a Pareto frontier. This could allow the electric utilities to trade-off battery lifetime and fuel efficiency at their preference while striking the high gain zone for remote microgrid fuel consumptions.



\subsubsection{Impact of the recovery shield}


\begin{figure*}[!t]
    \centering
    \begin{subfigure}{0.68\linewidth}
        \includegraphics[width=\linewidth]{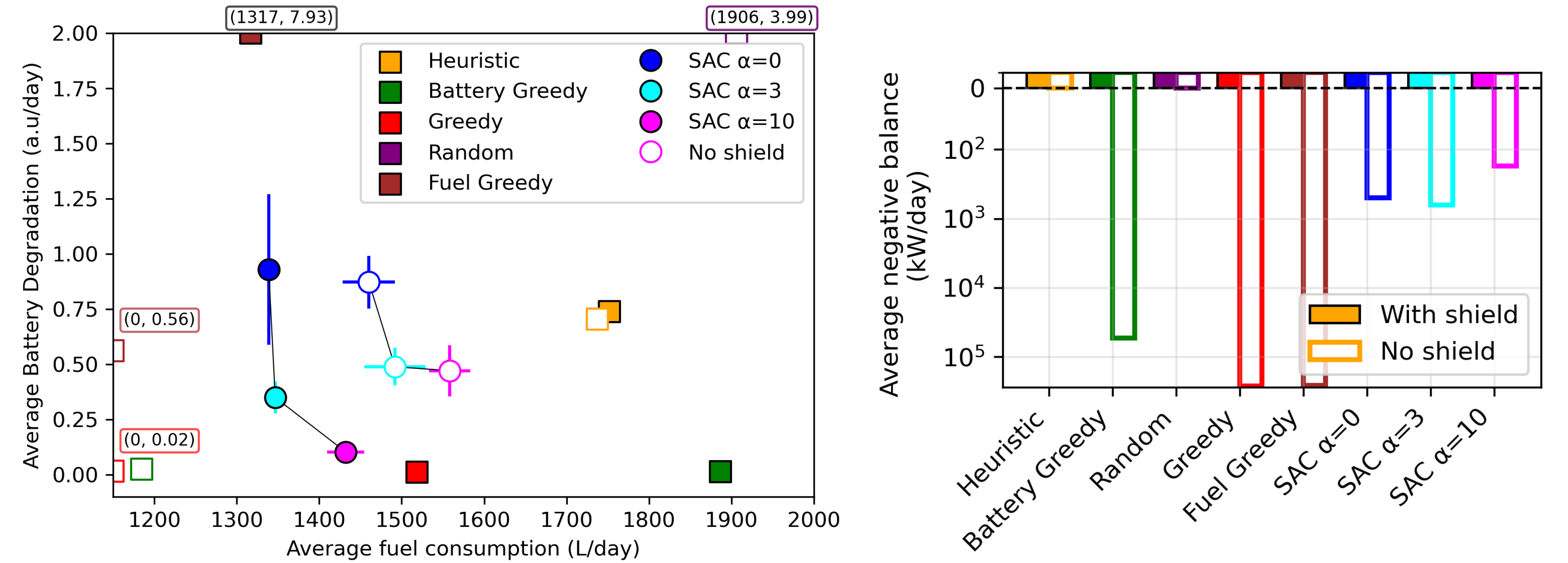}
        \caption{\em Left: Impact of the recovery shield. RL agents consume more fuel and degrade the battery more without shielding. Right: shielding prevents balance violations.}
        \label{fig:shielding_vs_no_shielding}
    \end{subfigure}
    \hfill
    \begin{subfigure}{0.31\linewidth}
        \includegraphics[width=\linewidth]{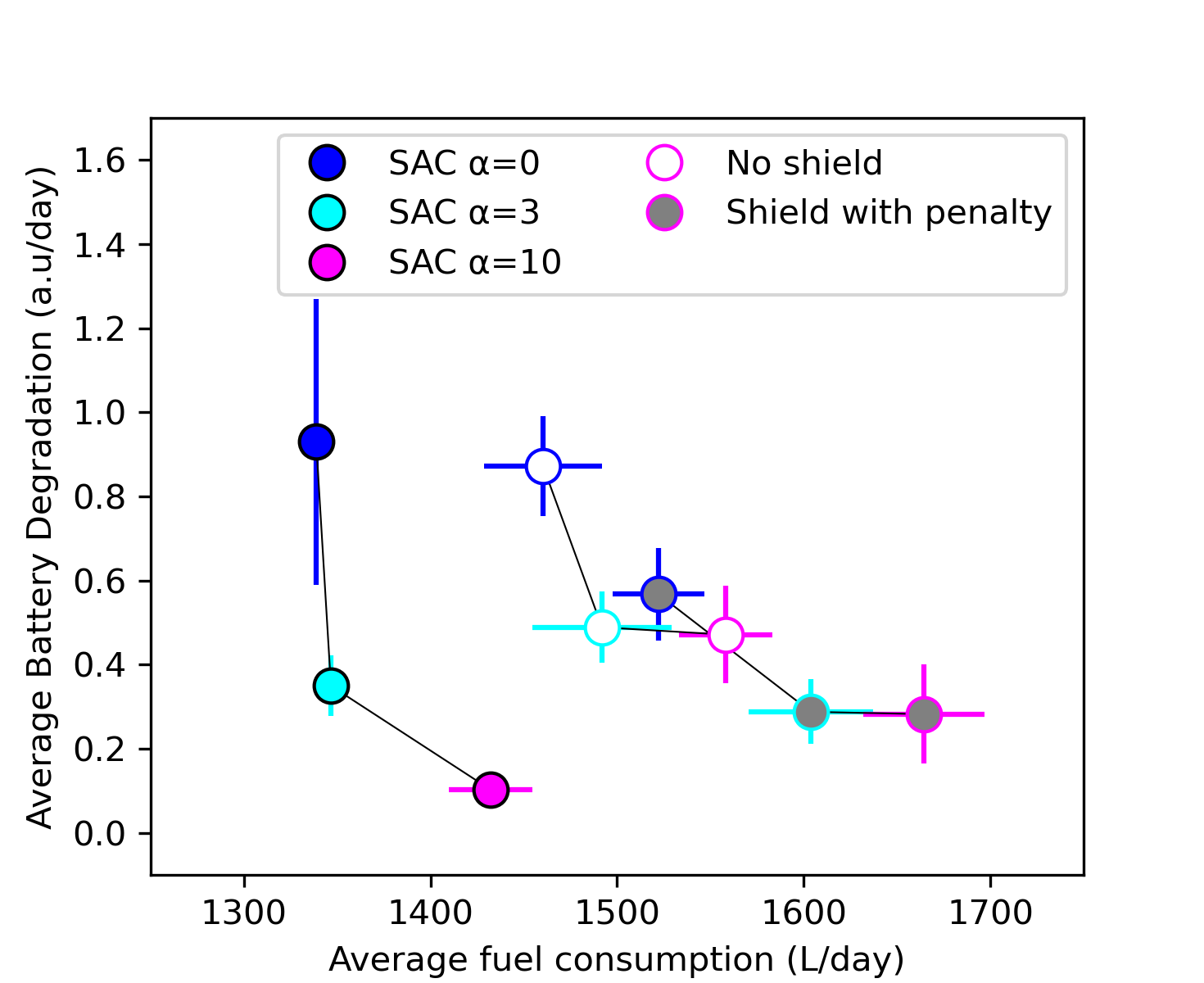}
        \caption{\em Shield penalty during training reduces performance. 
        }
        \label{fig:shied_usage_penalty}
    \end{subfigure}
    \caption{\em Performance of agents with and without recovery shielding in remote microgrid environments.}
    \label{fig:combined_rl_plots}
    \vspace{-5mm}
\end{figure*}

In this experiment, we explore the importance of recovery shielding for both RL agents and baseline algorithms. The recovery shield is the most complex shield in this system, addressing the uncertainty of exogenous variables and requiring multi-step simulation for compliance verification. To ensure compliance in every situation, the recovery shield considers worst-case scenario which could impair performance due to over restrictiveness. Figure~\ref{fig:shielding_vs_no_shielding} presents the performance metrics and negative balance violations, with and without recovery shield. Battery-greedy agent without recovery shield has a relatively good performance but it is not acceptable due to violating the negative balance constraint. Fuel-greedy agent without recovery shield does barely consume any fuel but it also show high amount of negative balance violation. On the other hand, Heuristic policy has a stable performance even without recovery shield which shows the validity of its policy. Finally, SAC agents trained and deployed without recovery shield, but penalized for negative balance violation exhibit increased fuel consumption and slightly higher battery degradation than their shielded counterparts, while still sometimes experiencing  negative balance (as shown in the right plot). We hypothesize that this occurs because they adopt an overly conservative and suboptimal policy due to the substantial penalties associated with negative balance during training. It is noteworthy that the RL agent with shielding shows much smaller standard errors in evaluated variables, demonstrating the robustness of training the RL agent with a shield.

\subsubsection{How to train an RL agent given a shield?}


A recurring research question is how penalizing the use of shielding during training impacts the agent's ability to learn an optimal policy~\citep{Hsu_Hu_Fisac_2023}. We explore this question for our use case by evaluating three variants of the RL agents in Figure~\ref{fig:shied_usage_penalty}: the first variant  (full color) was trained with recovery shielding but without penalizing the shield intervention. The second variant (gray) was trained with recovery shield and a penalty for shield intervention, and the third agent (white) was trained without recovery shield but with a penalty for violating the non-zero balance constraint. 

The agent trained with recovery shielding and no penalty (blue) achieves best performance. Both the orange and green agents learn to be more conservative due to the negative balance or shield penalty. Over-conservativeness is a known issue in punishing RL environments~\citep{mani2025risk}. Indeed, we notice these agents keep gensets \textit{on} longer which decreases the chance of receiving a penalty. Interestingly, the shield penalty seems to encourage the agent to charge the battery more probably because the shield penalty is harder to predict than the negative balance penalty for the agent.

\section{Discussion}

While remote microgrid optimization is complex, it remains at a sufficiently small scale that it can be analyzed in depth. \cite{Eichelbeck_Markgraf_Althoff_2022} have shown that, for a very similar problem, a provable, single shield can be hand-designed to respect constraints without significantly impairing the performance. However, their paper shows the challenge of designing such a shield to comply with all constraints at once.
Our approach's applicability stands out in comparison: for every SCU, the shield is simple as it only focuses on a specific set of constraints: the others are accounted for by the SCU structure. In most cases, fallback policies are evident. Handling uncertain dynamics, such as with exogenous variables for the microgrid SCU, can be managed with the necessary level of restrictiveness. In addition, SCUs allow to analyze the degrees of freedom that each constraints removes from the RL agent, and allow it to focus on these instead of outputting actions for every device. 

A potential limitation of the SCU approach is the duplication of lower-level digital twins inside higher-level digital twins. This allows higher-level shielded controllers to consider the lower-level constraints when using rolling out strategies, but could lead to an exponential number of digital twins depending on the number of SCU levels. However, in moderately complex real world cases, the number of SCU levels can be expected to be small \citep{siyari2019emergence}, such that scaling should not be a limiting factor.
As an example, in our case the SCU hierarchy has depth $3$, and each microgrid SCU step—including future projections—executes in under 0.05 seconds on a standard CPU, demonstrating that practical deployments remain well within real-time control.

\vspace{-3mm}
\section{Conclusion}
\label{sec:conclusion}

This paper presented Shielded Controller Units, a decomposition approach facilitating the design of white-box shields to ensure compliance for complex constrained systems controlled by RL agents. SCUs allow to design simple shields for every constraint, which are then combined hierarchically. Digital twins pertaining to each SCU are updated at every time step so higher-level units shields can rollout simulations accounting for lower-level constraints. 
SCUs were applied on a realistic remote microgrid optimization problem, ensuring the RL agent complies with every operational constraints. Every shield level was simple to design, describe and interpret, showcasing the relevance of SCU for provable RL compliance in complex real-world settings. 

By facilitating deployment of RL to critical systems requiring interpretable guarantees, SCUs can allow us to benefit from the potential of RL for optimizing decision-making with important social impact. We demonstrated this by showing how RL agents when supported by SCUs can reduce fuel consumption by 24\% for similar battery degradation compared to industry heuristics. We hope SCUs can be valuable for RL deployment in other regulated systems in the fields of energy, healthcare, transportation, etc.



\subsubsection*{Broader Impact Statement}
\label{sec:broaderImpact}
SCUs aim at facilitating the proof of constraint compliance for RL agents in complex systems and thus enhance RL applicability to critical industrial settings. This is an necessary step towards capitalizing on the potential of RL for real world applications. In the case of power systems such as microgrids, this could facilitate the integration of renewable energy sources towards the energy transition.   

\subsubsection*{Acknowledgments}
The authors would like to thank Mathieu Lambert for his precious advice, and Slavica Antic for her help gathering wind turbine data.

\bibliography{main}

\begin{thebibliography}{46}
\providecommand{\natexlab}[1]{#1}
\providecommand{\url}[1]{\texttt{#1}}
\expandafter\ifx\csname urlstyle\endcsname\relax
  \providecommand{\doi}[1]{DOI: #1}\else
  \providecommand{\doi}{DOI: \begingroup \urlstyle{rm}\Url}\fi

\bibitem[Achiam et~al.(2017)Achiam, Held, Tamar, and Abbeel]{achiam2017constrainedpolicyoptimization}
Joshua Achiam, David Held, Aviv Tamar, and Pieter Abbeel.
\newblock Constrained policy optimization, 2017.
\newblock URL \url{https://arxiv.org/abs/1705.10528}.

\bibitem[Alshiekh et~al.(2018)Alshiekh, Bloem, K{\"o}nighofer, Nimmermann, Schmuck, and Weinberger]{Alshiekh2018}
Mohammed Alshiekh, Roderick Bloem, Robert K{\"o}nighofer, Laura Nimmermann, Anna-Katharina Schmuck, and Stefan Weinberger.
\newblock Safe reinforcement learning via shielding.
\newblock In \emph{Proceedings of the AAAI Conference on Artificial Intelligence}, volume~32, 2018.

\bibitem[Amodei et~al.(2016)Amodei, Olah, Steinhardt, Christiano, Schulman, and Man{\'e}]{Amodei2016}
Dario Amodei, Chris Olah, Jacob Steinhardt, Paul Christiano, John Schulman, and Dan Man{\'e}.
\newblock Concrete problems in ai safety.
\newblock \emph{arXiv preprint arXiv:1606.06565}, 2016.

\bibitem[Arwa \& Folly(2020)Arwa and Folly]{Arwa2020}
E.~O. Arwa and K.~A. Folly.
\newblock Reinforcement learning techniques for optimal power control in grid-connected microgrids: A comprehensive review.
\newblock \emph{IEEE Access}, 8:\penalty0 208992--209007, 2020.
\newblock \doi{10.1109/ACCESS.2020.3038735}.

\bibitem[Banerjee et~al.(2024)Banerjee, Rahmani, Biswas, and Dillig]{Banerjee_Rahmani_Biswas_Dillig_2024}
Arko Banerjee, Kia Rahmani, Joydeep Biswas, and Isil Dillig.
\newblock Dynamic model predictive shielding for provably safe reinforcement learning.
\newblock \penalty0 (arXiv:2405.13863), Dec 2024.
\newblock URL \url{http://arxiv.org/abs/2405.13863}.
\newblock arXiv:2405.13863 [cs].

\bibitem[Bastani(2020)]{bastani2020safereinforcementlearningnonlinear}
Osbert Bastani.
\newblock Safe reinforcement learning with nonlinear dynamics via model predictive shielding, 2020.
\newblock URL \url{https://arxiv.org/abs/1905.10691}.

\bibitem[Bloem et~al.(2012)Bloem, Jobstmann, Piterman, Pnueli, and Sa'ar]{bloem2012synthesis}
Roderick Bloem, Barbara Jobstmann, Nir Piterman, Amir Pnueli, and Yaniv Sa'ar.
\newblock Synthesis of reactive (1) designs.
\newblock \emph{Journal of Computer and System Sciences}, 78\penalty0 (3):\penalty0 911--938, 2012.

\bibitem[Bloem et~al.(2020)Bloem, Jensen, Könighofer, Larsen, Lorber, and Palmisano]{Bloem_2020}
Roderick Bloem, Peter~Gjøl Jensen, Bettina Könighofer, Kim~Guldstrand Larsen, Florian Lorber, and Alexander Palmisano.
\newblock It’s time to play safe: Shield synthesis for timed systems.
\newblock \penalty0 (arXiv:2006.16688), Jun 2020.
\newblock \doi{10.48550/arXiv.2006.16688}.
\newblock URL \url{http://arxiv.org/abs/2006.16688}.
\newblock arXiv:2006.16688 [cs].

\bibitem[Bouton et~al.(2019)Bouton, Karlsson, Nakhaei, Fujimura, Kochenderfer, and Tumova]{Bouton_Karlsson_Nakhaei_Fujimura_Kochenderfer_Tumova_2019}
Maxime Bouton, Jesper Karlsson, Alireza Nakhaei, Kikuo Fujimura, Mykel~J. Kochenderfer, and Jana Tumova.
\newblock Reinforcement learning with probabilistic guarantees for autonomous driving.
\newblock \penalty0 (arXiv:1904.07189), May 2019.
\newblock \doi{10.48550/arXiv.1904.07189}.
\newblock URL \url{http://arxiv.org/abs/1904.07189}.
\newblock arXiv:1904.07189 [cs].

\bibitem[Cao et~al.(2020)Cao, Harrold, Fan, Morstyn, Healey, and Li]{Cao_2020}
Jun Cao, Dan Harrold, Zhong Fan, Thomas Morstyn, David Healey, and Kang Li.
\newblock Deep reinforcement learning-based energy storage arbitrage with accurate lithium-ion battery degradation model.
\newblock \emph{IEEE Transactions on Smart Grid}, 11\penalty0 (5):\penalty0 4513–4521, September 2020.
\newblock ISSN 1949-3053, 1949-3061.
\newblock \doi{10.1109/TSG.2020.2986333}.

\bibitem[Carr et~al.(2022)Carr, Jansen, Junges, and Topcu]{Carr_Jansen_Junges_Topcu_2022}
Steven Carr, Nils Jansen, Sebastian Junges, and Ufuk Topcu.
\newblock Safe reinforcement learning via shielding under partial observability.
\newblock \penalty0 (arXiv:2204.00755), Aug 2022.
\newblock \doi{10.48550/arXiv.2204.00755}.
\newblock URL \url{http://arxiv.org/abs/2204.00755}.
\newblock arXiv:2204.00755 [cs].

\bibitem[Ceusters et~al.(2023)Ceusters, Camargo, Franke, Nowé, and Messagie]{Ceusters_Camargo_Franke_Nowé_Messagie_2023}
Glenn Ceusters, Luis~Ramirez Camargo, Rüdiger Franke, Ann Nowé, and Maarten Messagie.
\newblock Safe reinforcement learning for multi-energy management systems with known constraint functions.
\newblock \emph{Energy and AI}, 12:\penalty0 100227, Apr 2023.
\newblock ISSN 2666-5468.
\newblock \doi{10.1016/j.egyai.2022.100227}.

\bibitem[Chen et~al.(2021)Chen, Shi, Arnold, and Peisert]{Chen_Shi_Arnold_Peisert_2021}
Yize Chen, Yuanyuan Shi, Daniel Arnold, and Sean Peisert.
\newblock Saver: Safe learning-based controller for real-time voltage regulation.
\newblock \penalty0 (arXiv:2111.15152), Nov 2021.
\newblock \doi{10.48550/arXiv.2111.15152}.
\newblock URL \url{http://arxiv.org/abs/2111.15152}.
\newblock arXiv:2111.15152 [cs, eess].

\bibitem[David et~al.(2023)David, Massé, and Zinflou]{demand-forecast}
Charline David, Alexandre~Blondin Massé, and Arnaud Zinflou.
\newblock Fast short-term electrical load forecasting under high meteorological variability with a multiple equation time series approach.
\newblock \emph{International Journal of Electrical and Computer Engineering}, 17\penalty0 (10):\penalty0 234 -- 241, 2023.
\newblock ISSN eISSN: 1307-6892.
\newblock URL \url{https://publications.waset.org/vol/202}.

\bibitem[Dimeas \& Hatziargyriou(2010)Dimeas and Hatziargyriou]{Dimeas2010}
A.~L. Dimeas and N.~D. Hatziargyriou.
\newblock Multi-agent reinforcement learning for microgrids.
\newblock In \emph{IEEE PES General Meeting}, pp.\  1--8, July 2010.
\newblock \doi{10.1109/PES.2010.5589633}.

\bibitem[Dulac-Arnold et~al.(2019)Dulac-Arnold, Mankowitz, and Hester]{Dulac-Arnold2019}
Gabriel Dulac-Arnold, Daniel Mankowitz, and Todd Hester.
\newblock Challenges of real-world reinforcement learning.
\newblock \emph{arXiv preprint arXiv:1904.12901}, 2019.

\bibitem[Eichelbeck et~al.(2022)Eichelbeck, Markgraf, and Althoff]{Eichelbeck_Markgraf_Althoff_2022}
Michael Eichelbeck, Hannah Markgraf, and Matthias Althoff.
\newblock Contingency-constrained economic dispatch with safe reinforcement learning.
\newblock In \emph{2022 21st IEEE International Conference on Machine Learning and Applications (ICMLA)}, pp.\  597–602, Dec 2022.
\newblock \doi{10.1109/ICMLA55696.2022.00103}.
\newblock URL \url{http://arxiv.org/abs/2205.06212}.
\newblock arXiv:2205.06212 [cs, eess].

\bibitem[Foruzan et~al.(2018)Foruzan, Soh, and Asgarpoor]{Foruzan2018}
E.~Foruzan, L.-K. Soh, and S.~Asgarpoor.
\newblock Reinforcement learning approach for optimal distributed energy management in a microgrid.
\newblock \emph{IEEE Transactions on Power Systems}, 33\penalty0 (5):\penalty0 5749--5758, September 2018.
\newblock \doi{10.1109/TPWRS.2018.2823641}.

\bibitem[Fulton \& Platzer(2018)Fulton and Platzer]{Fulton_Platzer_2018}
Nathan Fulton and André Platzer.
\newblock Safe reinforcement learning via formal methods: Toward safe control through proof and learning.
\newblock \emph{Proceedings of the AAAI Conference on Artificial Intelligence}, 32\penalty0 (11), Apr 2018.
\newblock ISSN 2374-3468.
\newblock \doi{10.1609/aaai.v32i1.12107}.
\newblock URL \url{https://ojs.aaai.org/index.php/AAAI/article/view/12107}.

\bibitem[Garc{\'\i}a \& Fern{\'a}ndez(2015)Garc{\'\i}a and Fern{\'a}ndez]{Garcia2015}
Javier Garc{\'\i}a and Fernando Fern{\'a}ndez.
\newblock A comprehensive survey on safe reinforcement learning.
\newblock \emph{Journal of Machine Learning Research}, 16\penalty0 (1):\penalty0 1437--1480, 2015.

\bibitem[Haarnoja et~al.(2018)Haarnoja, Zhou, Abbeel, and Levine]{haarnoja18}
Tuomas Haarnoja, Aurick Zhou, Pieter Abbeel, and Sergey Levine.
\newblock Soft actor-critic: Off-policy maximum entropy deep reinforcement learning with a stochastic actor.
\newblock In Jennifer Dy and Andreas Krause (eds.), \emph{Proceedings of the 35th International Conference on Machine Learning}, volume~80 of \emph{Proceedings of Machine Learning Research}, pp.\  1861--1870. PMLR, 10--15 Jul 2018.
\newblock URL \url{https://proceedings.mlr.press/v80/haarnoja18b.html}.

\bibitem[Hajimiragha \& Zadeh(2013)Hajimiragha and Zadeh]{Hajimiragha2013}
A.~H. Hajimiragha and M.~R.~D. Zadeh.
\newblock Research and development of a microgrid control and monitoring system for the remote community of bella coola: Challenges, solutions, achievements and lessons learned.
\newblock In \emph{2013 IEEE International Conference on Smart Energy Grid Engineering (SEGE)}, pp.\  1--6, August 2013.
\newblock \doi{10.1109/SEGE.2013.6707898}.

\bibitem[Hassani \& Lambert(2024)Hassani and Lambert]{G-2024-49}
Rachid Hassani and Mathieu Lambert.
\newblock Real-time battery optimization for remote microgrids.
\newblock {Les Cahiers du GERAD} G-2024-49, Groupe d’études et de recherche en analyse des décisions, GERAD, Montréal QC H3T 2A7, Canada, August 2024.

\bibitem[Hirwa et~al.(2022)Hirwa, Ogunmodede, Zolan, and Newman]{Hirwa2022}
J.~Hirwa, O.~Ogunmodede, A.~Zolan, and A.~M. Newman.
\newblock Optimizing design and dispatch of a renewable energy system with combined heat and power.
\newblock \emph{Optimization and Engineering}, 23\penalty0 (3):\penalty0 1--31, September 2022.
\newblock \doi{10.1007/s11081-021-09674-4}.

\bibitem[Hsu et~al.(2023)Hsu, Hu, and Fisac]{Hsu_Hu_Fisac_2023}
Kai-Chieh Hsu, Haimin Hu, and Jaime~Fernández Fisac.
\newblock The safety filter: A unified view of safety-critical control in autonomous systems.
\newblock \penalty0 (arXiv:2309.05837), Sep 2023.
\newblock \doi{10.48550/arXiv.2309.05837}.
\newblock URL \url{http://arxiv.org/abs/2309.05837}.
\newblock arXiv:2309.05837 [cs, eess].

\bibitem[{International Organization for Standardization (ISO)}(2018)]{iso8528-1-2018}
{International Organization for Standardization (ISO)}.
\newblock Reciprocating internal combustion engine driven alternating current generating sets — part 1: Application, ratings and performance, 2018.
\newblock URL \url{https://www.iso.org/standard/68539.html}.
\newblock Edition 3.

\bibitem[Jansen et~al.(2018)Jansen, K{\"o}nighofer, Junges, and Bloem]{Jansen2018}
Nils Jansen, Bettina K{\"o}nighofer, Sebastian Junges, and Roderick Bloem.
\newblock Shielded decision-making in mdps.
\newblock In \emph{arXiv preprint arXiv:1807.06096}, 2018.

\bibitem[Kober et~al.(2013)Kober, Bagnell, and Peters]{Kober2013}
Jens Kober, J~Andrew Bagnell, and Jan Peters.
\newblock Reinforcement learning in robotics: A survey.
\newblock \emph{The International Journal of Robotics Research}, 32\penalty0 (11):\penalty0 1238--1274, 2013.

\bibitem[K{\"o}nighofer et~al.(2020)K{\"o}nighofer, Alshiekh, Bloem, Humphrey, K{\"o}nighofer, Topcu, and Wang]{Könighofer2020}
Bettina K{\"o}nighofer, Mohammed Alshiekh, Roderick Bloem, Laura Humphrey, Robert K{\"o}nighofer, Ufuk Topcu, and Chao Wang.
\newblock Shield synthesis.
\newblock \emph{Formal Methods in System Design}, 57:\penalty0 79--115, 2020.

\bibitem[Krasowski et~al.(2023)Krasowski, Thumm, Müller, Schäfer, Wang, and Althoff]{Krasowski_Thumm_Muller_Schafer_Wang_Althoff_2023}
Hanna Krasowski, Jakob Thumm, Marlon Müller, Lukas Schäfer, Xiao Wang, and Matthias Althoff.
\newblock Provably safe reinforcement learning: Conceptual analysis, survey, and benchmarking.
\newblock \penalty0 (arXiv:2205.06750), Nov 2023.
\newblock \doi{10.48550/arXiv.2205.06750}.
\newblock URL \url{http://arxiv.org/abs/2205.06750}.
\newblock arXiv:2205.06750 [cs].

\bibitem[Kwon \& Zhu(2022)Kwon and Zhu]{Kwon_Zhu_2022}
Kyung-Bin Kwon and Hao Zhu.
\newblock Reinforcement learning-based optimal battery control under cycle-based degradation cost.
\newblock \emph{IEEE Transactions on Smart Grid}, 13\penalty0 (6):\penalty0 4909–4917, Nov 2022.
\newblock ISSN 1949-3053, 1949-3061.
\newblock \doi{10.1109/TSG.2022.3180674}.

\bibitem[Lambert \& Hassani(2023)Lambert and Hassani]{Lambert2023}
M.~Lambert and R.~Hassani.
\newblock Diesel genset optimization in remote microgrids.
\newblock \emph{Applied Energy}, 340:\penalty0 121036, June 2023.
\newblock \doi{10.1016/j.apenergy.2023.121036}.

\bibitem[Liu et~al.(2021)Liu, Halev, and Liu]{ijcai2021p614}
Yongshuai Liu, Avishai Halev, and Xin Liu.
\newblock Policy learning with constraints in model-free reinforcement learning: A survey.
\newblock In Zhi-Hua Zhou (ed.), \emph{Proceedings of the Thirtieth International Joint Conference on Artificial Intelligence, {IJCAI-21}}, pp.\  4508--4515. International Joint Conferences on Artificial Intelligence Organization, 8 2021.
\newblock \doi{10.24963/ijcai.2021/614}.
\newblock URL \url{https://doi.org/10.24963/ijcai.2021/614}.
\newblock Survey Track.

\bibitem[Liu et~al.(2022)Liu, Cen, Isenbaev, Liu, Wu, Li, and Zhao]{liu2022constrainedvariationalpolicyoptimization}
Zuxin Liu, Zhepeng Cen, Vladislav Isenbaev, Wei Liu, Zhiwei~Steven Wu, Bo~Li, and Ding Zhao.
\newblock Constrained variational policy optimization for safe reinforcement learning, 2022.
\newblock URL \url{https://arxiv.org/abs/2201.11927}.

\bibitem[Mani et~al.(2025)Mani, Mai, Gauthier, Chen, Nashed, and Paull]{mani2025risk}
Kaustubh Mani, Vincent Mai, Charlie Gauthier, Annie~S Chen, Samer~B. Nashed, and Liam Paull.
\newblock Risk informed policy learning for safer exploration.
\newblock In \emph{The Thirteenth International Conference on Learning Representations}, 2025.
\newblock URL \url{https://openreview.net/forum?id=gJG4IPwg6l}.

\bibitem[Obermayr et~al.(2021)Obermayr, Riess, and Wilde]{Obermayr_Riess_Wilde_2021}
Martin Obermayr, Christian Riess, and Jürgen Wilde.
\newblock A novel online 4-point rainflow counting algorithm for power electronics.
\newblock \emph{Microelectronics Reliability}, 120:\penalty0 114112, May 2021.
\newblock ISSN 00262714.
\newblock \doi{10.1016/j.microrel.2021.114112}.

\bibitem[Schulman et~al.(2017)Schulman, Wolski, Dhariwal, Radford, and Klimov]{schulman2017proximalpolicyoptimizationalgorithms}
John Schulman, Filip Wolski, Prafulla Dhariwal, Alec Radford, and Oleg Klimov.
\newblock Proximal policy optimization algorithms, 2017.
\newblock URL \url{https://arxiv.org/abs/1707.06347}.

\bibitem[Silvente et~al.(2015)Silvente, Kopanos, Pistikopoulos, and Espuña]{Silvente2015}
J.~Silvente, G.~M. Kopanos, E.~N. Pistikopoulos, and A.~Espuña.
\newblock A rolling horizon optimization framework for the simultaneous energy supply and demand planning in microgrids.
\newblock \emph{Applied Energy}, 155:\penalty0 485--501, October 2015.
\newblock \doi{10.1016/j.apenergy.2015.05.090}.

\bibitem[Siyari et~al.(2019)Siyari, Dilkina, and Dovrolis]{siyari2019emergence}
Payam Siyari, Bistra Dilkina, and Constantine Dovrolis.
\newblock Emergence and evolution of hierarchical structure in complex systems.
\newblock In \emph{Dynamics On and Of Complex Networks III: Machine Learning and Statistical Physics Approaches 10}, pp.\  23--62. Springer, 2019.

\bibitem[Wachi et~al.(2024)Wachi, Shen, and Sui]{wachi2024surveyconstraintformulationssafe}
Akifumi Wachi, Xun Shen, and Yanan Sui.
\newblock A survey of constraint formulations in safe reinforcement learning, 2024.
\newblock URL \url{https://arxiv.org/abs/2402.02025}.

\bibitem[Waga et~al.(2022)Waga, Castellano, Pruekprasert, Klikovits, Takisaka, and Hasuo]{waga2022dynamicshieldingreinforcementlearning}
Masaki Waga, Ezequiel Castellano, Sasinee Pruekprasert, Stefan Klikovits, Toru Takisaka, and Ichiro Hasuo.
\newblock Dynamic shielding for reinforcement learning in black-box environments, 2022.
\newblock URL \url{https://arxiv.org/abs/2207.13446}.

\bibitem[Wang et~al.(2024)Wang, Xiao, You, and Poor]{Wang2024}
Y.~Wang, M.~Xiao, Y.~You, and H.~V. Poor.
\newblock Optimized energy dispatch for microgrids with distributed reinforcement learning.
\newblock \emph{IEEE Transactions on Smart Grid}, pp.\  1--1, 2024.
\newblock \doi{10.1109/TSG.2023.3331467}.

\bibitem[Xu et~al.(2018)Xu, Oudalov, Ulbig, Andersson, and Kirschen]{Xu_Oudalov_Ulbig_Andersson_Kirschen_2018}
Bolun Xu, Alexandre Oudalov, Andreas Ulbig, Goran Andersson, and Daniel~S. Kirschen.
\newblock Modeling of lithium-ion battery degradation for cell life assessment.
\newblock \emph{IEEE Transactions on Smart Grid}, 9\penalty0 (2):\penalty0 1131–1140, Mar 2018.
\newblock ISSN 1949-3053, 1949-3061.
\newblock \doi{10.1109/TSG.2016.2578950}.

\bibitem[Yang et~al.(2021)Yang, Zhao, Li, and Zomaya]{Yang2021}
T.~Yang, L.~Zhao, W.~Li, and A.~Y. Zomaya.
\newblock Dynamic energy dispatch strategy for integrated energy system based on improved deep reinforcement learning.
\newblock \emph{Energy}, 235:\penalty0 121377, November 2021.
\newblock \doi{10.1016/j.energy.2021.121377}.

\bibitem[Yu et~al.(2024)Yu, Zhang, Song, Hui, and Chen]{Yu_Zhang_Song_Hui_Chen_2024}
Peipei Yu, Hongcai Zhang, Yonghua Song, Hongxun Hui, and Ge~Chen.
\newblock District cooling system control for providing operating reserve based on safe deep reinforcement learning.
\newblock \emph{IEEE Transactions on Power Systems}, 39\penalty0 (1):\penalty0 40–52, Jan 2024.
\newblock ISSN 0885-8950, 1558-0679.
\newblock \doi{10.1109/TPWRS.2023.3237888}.

\bibitem[Zhang et~al.(2024)Zhang, Guan, Che, and Shahidehpour]{Zhang_Guan_Che_Shahidehpour_2024}
Jin Zhang, Yuxiang Guan, Liang Che, and Mohammad Shahidehpour.
\newblock Ev charging command fast allocation approach based on deep reinforcement learning with safety modules.
\newblock \emph{IEEE Transactions on Smart Grid}, 15\penalty0 (1):\penalty0 757–769, Jan 2024.
\newblock ISSN 1949-3053, 1949-3061.
\newblock \doi{10.1109/TSG.2023.3281782}.

\end{thebibliography}
\bibliographystyle{rlj}

\newpage

\beginSupplementaryMaterials

\appendix

\section{Agent architecture}
\label{app:agent_architecture}

\begin{figure}[h]
    \centering
    
    \includegraphics[width=0.75\linewidth]{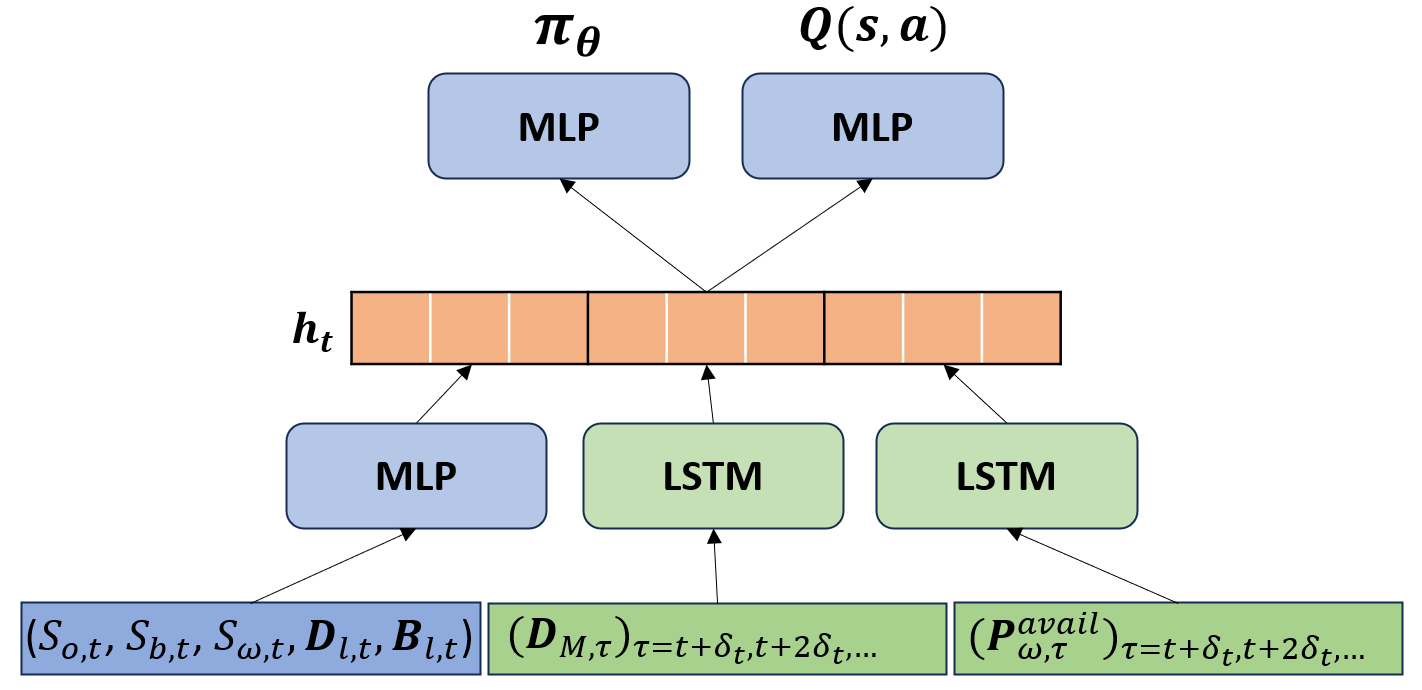}

    \caption{The agent architecture}
    \label{fig:agent_arch}
\end{figure}

All RL agents use SAC for decision making. Both the actor and the critic use the same architecture (See Figure~\ref{fig:agent_arch}). The state is formatted as a vector and fed into a multi-layer perceptron (MLP). To take into account the predicted future demand and wind power, we added two LSTM modules which takes the demand and wind power predictions and concatenate their hidden representations with the encoded representation of the state. The final representation is passed through another MLP to compute the action probabilities and state values.

\section{Learning hyper-parameters}
\label{app:learning_params} 

In Table~\ref{tab:learning_params}, we provide the values for the learning agent and the optimization hyper-parameters. 

\begin{table}[h]
    \centering
    \begin{tabular}{|c|c|c|}
    \hline
       \multicolumn{3}{|c|}{\textbf{SAC agent}}\\
       \hline
       Observation encoder & MLP  &  2 layers (128, 32) \\       
       \hline
       Hidden dimension size & $H$ & $3\times 32$\\
       \hline
    \multicolumn{3}{|c|}{\textbf{Optimization}}\\
      \hline      
       Learning Rate & $lr$  & 0.001, \textbf{0.0003} \\ 
      \hline
      Batch size & $B$ & \textbf{32}, 64 \\      \hline
      
    \hline
    \end{tabular}
    \caption{Learning Hyper-parameters. The final selected values are in bold.}
    \label{tab:learning_params}
\end{table}

\section{Battery Degradation}\label{app:battery_degradation}

Different factors can degrade a battery and reduce its storage capacity. Time and temperature affect its capacity, however, these are not under the agent's control. More relevant here, cycling, i.e. repeated charging and discharging, significantly impacts  the battery's degradation \citep{Cao_2020}. Cycle-based battery degradation is often estimated through the rainflow analysis of the battery's state of charge  (SoC) \citep{Xu_Oudalov_Ulbig_Andersson_Kirschen_2018}, which accounts for the number and amplitude of cycles along time. However, this process is usually done offline, given a history of data. In RL however, a degradation cost $d_{b,t}$ at every time step is needed that reflects the impact of the agent's action while, when summed over a trajectory, being equal to the result of the offline process.

A commonly used approach for online battery degradation estimation is to use a linear approximation, as presented by \cite{Cao_2020}.  The linearized degradation cost is calculated using a simplified linear model, where the cost is directly proportional to the absolute value of the battery SoC change, $\text{b}_t = \text{soc}_{t} - \text{soc}_{t-1}$. The linearized degradation cost is thus computed as $d_{b,t} = \alpha_d \cdot |\text{b}_t| $ where $\alpha_d$ is the degradation coefficient. This linear approach provides a straightforward estimation of the degradation cost, making it easier to implement and understand while still capturing the essential impact of battery usage on its lifespan.  

In this paper, we aimed at representing the industrial environment as precisely as possible, and as such, to model the battery degradation more accurately. \cite{Kwon_Zhu_2022} have proposed an approach to online cycle-based degradation cost estimation for RL. In battery fatigue modeling, the impact of a cycle on degradation is exponential to the cycle amplitude. An interesting insight from \cite{Kwon_Zhu_2022} is their decomposition of the cost for each time step as
$$ d_{b,t} = \alpha_d \cdot \exp(\beta \cdot |\text{soc}_t + b_t - R[-1]|) - \alpha_d \cdot \exp(\beta \cdot |\text{soc}_t - R[-1]|)$$
where $\alpha_d$ represents the degradation rate, $\beta$ indicates the sensitivity to SoC changes, and $ R[-1] $ is the most recent switching point in the SoC history. 

The difficulty of the rainflow algorithm is however to find the right value of $R[-1]$, as cycles can close, and at the next time step, the SoC evolution can pertain to another cycle open further in history. An important problem is that, in theory, the necessary memory to correctly assign each time step to the correct value of $R[-1]$ can be as long as the complete SoC history. \cite{Kwon_Zhu_2022} propose an algorithm to keep only three points, however, it does not fully respect rainflow analysis. Instead, we based our approach on \cite{Obermayr_Riess_Wilde_2021}, where a 4-point online rainflow algorithm allows to identify the last switching point while memory requirement is hard-capped thanks to a discretization process. 
Our slightly adapted algorithm is summarized in Algorithms \ref{alg:Rainflow4P} and \ref{alg:HysteresisFilter}. The resulting buffer $R$ is used to compute the reward, and is given as an observation to the agent as part of the battery state to respect the Markovian assumption for the environment.

Note that the discretization can cause artifacts, such as the calculated degradation cost $d_{b,t}$ being negative close to switching points. In this case, we instead approximate it for this time step based on the discretization window $w$: $d_{b,t} = \alpha_d \cdot (\exp(\beta \cdot |w|) - 1) / w$.

A final question regards the value of $\alpha_d$ and $\beta$, so that $d_{b,t}$ represents a realistic capacity loss for the battery. We chose $\beta = 1$ and $\alpha_d = 5$ as in \citep{Cao_2020}. However, these values should be calibrated with a real battery if to be deployed in the real world. This explains our choice to describe the battery degradation units are arbitrary in the paper, and keep the reward function flexible with different values of weight $\alpha$.

This approach ensures that the degradation cost accurately reflects the impact of battery usage on its lifespan, considering both the magnitude and frequency of usage changes. 

\newpage

\begin{algorithm}[H]
\caption{Rainflow 4-point algorithm}\label{alg:Rainflow4P}
\begin{algorithmic}
\State \textcolor[rgb]{0.0, 0.5, 0.0}{\textit{// 4-point rainflow condition check}}
\Function{Rainflow4P}{R}
    \State $cycleFound \gets False$
    \If{$\min(R[-4], R[-1]) \leq \min(R[-3], R[-2])$ \textbf{and} $\max(R[-3], R[-2]) \leq \max(R[-4], R[-1])$}
        \State $cycleFound \gets True$
    \EndIf
    \State \Return $cycleFound$
\EndFunction

\State \textcolor[rgb]{0.0, 0.5, 0.0}{\textit{// Main function to update switching point buffer R}}
\Function{UpdateSwitchingPoints}{x, F, R}
    \State $F[2] \gets \textbf{Discretize}(x, d\_load)$
    \State $F, tpFound \gets \textbf{HysteresisFilter}(F)$
    \If{$tpFound$}
        \State \textbf{Append}(R, \textbf{Discretize}(F[0], d\_load))
    \EndIf
    \State \textbf{Append}(R, \textbf{Discretize}(x, d\_load))

    \While{$\textbf{Length}(R) \geq 4$ \textbf{and} \textbf{Rainflow4P}(R)}
        \State $R[-3] \gets R[-1]$
        \State \textbf{Delete}(R[-2:])
    \EndWhile
    \State \textbf{Delete}(R[-1])
    \State \Return $F, R$
\EndFunction

\end{algorithmic}
\end{algorithm}
\begin{algorithm}[H]
\caption{Hysteresis Filter algorithm}\label{alg:HysteresisFilter}
\begin{algorithmic}

\Function{HysteresisFilter}{F}
    \State \textcolor[rgb]{0.0, 0.5, 0.0}{\textit{// Apply hysteresis filter to the input signal}}
    \State \textcolor[rgb]{0.0, 0.5, 0.0}{\textit{// Define helper functions}}
    \Function{Shift}{F}
        \State $F[0] \gets F[1]$
        \State $F[1] \gets F[2]$
        \State \Return $F$
    \EndFunction
    \Function{Skip}{F}
        \State $F[1] \gets F[2]$
        \State \Return $F$
    \EndFunction

    \State $tpFound \gets False$
    \If{$F[2] < F[1]$}
        \If{$F[0] \geq F[1]$}
            \State $F \gets \textbf{Skip}(F)$
        \Else
            \State $tpFound \gets True$
            \State $F \gets \textbf{Shift}(F)$
        \EndIf
    \ElsIf{$F[2] > F[1]$}
        \If{$F[0] \leq F[1]$}
            \State $F \gets \textbf{Skip}(F)$
        \Else
            \State $tpFound \gets True$
            \State $F \gets \textbf{Shift}(F)$
        \EndIf
    \Else
        \If{$F[0] > F[1]$}
            \State $F[1] \gets \min(F[1], F[2])$
        \Else
            \State $F[1] \gets \max(F[1], F[2])$
        \EndIf
    \EndIf
    \State \Return $F, tpFound$
\EndFunction

\end{algorithmic}
\end{algorithm}

\newpage

\end{document}